\documentclass[nohyperref]{article}

\usepackage{microtype}
\usepackage{graphicx}
\usepackage{subfigure}
\usepackage{booktabs} 

\usepackage{hyperref}

\usepackage[accepted]{icml2023}

\usepackage{amsmath}
\usepackage{amssymb}
\usepackage{mathtools}
\usepackage{amsthm}
\usepackage{float}
\usepackage{caption}

\usepackage[capitalize,noabbrev]{cleveref}

\theoremstyle{plain}

\theoremstyle{definition}

\theoremstyle{remark}

\usepackage[textsize=tiny]{todonotes}

\icmltitlerunning{Lifting Discrete Diffusion to the Probability Simplex}

\begin{document}

\twocolumn[
\icmltitle{Diffusion on the Probability Simplex}



\icmlsetsymbol{equal}{*}

\begin{icmlauthorlist}
\icmlauthor{Griffin Floto}{eai}
\icmlauthor{Thorsteinn Jonsson}{eai}
\icmlauthor{Mihai Nica}{guelph}
\icmlauthor{Scott Sanner}{uoft}
\icmlauthor{Eric Zhengyu Zhu}{uoft}
\end{icmlauthorlist}

\icmlaffiliation{uoft}{Department of Computer Science, University of Toronto}
\icmlaffiliation{eai}{EthicalAI}
\icmlaffiliation{guelph}{Department of Mathematics and Statistics, University of Guelph}

\icmlcorrespondingauthor{Griffin Floto}{griffin@ethicalairesearch.com}

\icmlkeywords{Machine Learning, ICML}

\vskip 0.3in
]



\printAffiliationsAndNotice{\icmlEqualContribution} 

\begin{abstract}
Diffusion models learn to reverse the progressive noising of a data distribution to create a generative model. However, the desired continuous nature of the noising process can be at odds with discrete data. To deal with this tension between continuous and discrete objects, we propose a method of performing diffusion on the probability simplex. Using the probability simplex naturally creates an interpretation where points correspond to categorical probability distributions. Our method uses the softmax function applied to an Ornstein-Unlenbeck Process, a well-known stochastic differential equation. We find that our methodology also naturally extends to include diffusion on the unit cube which has applications for bounded image generation.
\end{abstract}

\section{Introduction}
Diffusion models \cite{sohldickstein2015deep} \cite{diff_ho} \cite{song_grad_diff} have emerged as a well-established class of generative models, finding applications in image \cite{dhariwal2021diffusion}, speech \cite{jeong2021difftts}, and video \cite{singer2022makeavideo} domains. Diffusion processes work by progressively adding noise to data, which transforms a complex data distribution into a simpler, easy-to-sample distribution. Diffusion models are used to reverse the noising process by learning a stochastic differential equation (SDE) parameterized by a neural network that generates the data distribution \cite{song2021scorebased}. 

In comparison to other popular methods, such as Generative Adversarial Networks \cite{gan}, diffusion models present a compelling advantage as they have an exact likelihood interpretation and do not require adversarial training that other state-of-the-art generative models require. That is, diffusion models enjoy the benefit of having a more stable training process that avoid non-overlapping data and generated distributions \citep{yang2023diffusion}. Furthermore, diffusion models are also advantageous over discretized normalizing flows, which face practical restrictions when computing the determinant of the Jacobian from the change of variables formula \cite{neural_ode}. 

Most work with diffusion models assume a continuous data distribution in $\mathbb{R}^n$ and noising is performed with Gaussian distributions. This presents a problem for discrete sampling: how would one add continuous Gaussian noise if the underlying categories are discrete? We propose the simple solution to perform diffusion by sampling from $k$ categories on the \emph{probability simplex} $\mathcal{S}^k := \{\mathbf{x}\in \mathbb{R}^k: 0\leq \mathbf{x}_i \leq 1, \sum_{i=1}^k \mathbf{x}_i = 1\}$. The result of the diffusion is interpreted as the \emph{probability} that a given category is chosen. By shifting from categories themselves, to the space of probabilities over categories, we effectively turn a discrete problem into a continuous one.

\section{Background}
\subsection{Diffusion with Score-Matching}
Score matching as formulated by \cite{song2021scorebased} considers a continuous time diffusion process. Typically, the forward process does not have parameters and is independent of the data distribution. In particular, the forward process is described by an SDE 

\begin{equation} \label{eqn:sde-fwd-process}
    \mathrm{d}\mathbf{x}_t = \mathbf{f}(\mathbf{x}_t,t)\mathrm{d}t + \mathbf{G}(\mathbf{x}_t,t)\mathrm{d}\mathbf{w}_t
\end{equation}

where $\mathbf{w}$ is the standard Wiener process (also know as Brownian motion), $\mathbf{f}(\cdot,t) : \mathbb{R}^d \rightarrow \mathbb{R}^d$ is the drift term and $\mathbf{G}(\cdot,t) : \mathbb{R}^d \rightarrow \mathbb{R}^{d \times d}$ is the diffusion coefficient. The process maps a data distribution, $p_{t=0}(\mathbf{x}_t) \in \mathbb{R}^d$ into some limiting distribution $p_{t=1}(\mathbf{x}_t)$. The limiting distribution is chosen to be easy to sample from, and independent from the data distribution. Classical results in the theory of stochastic processes then tell us that the time reverse of this process is itself an SDE and obeys

\begin{equation} \label{eqn:sde-rev-process}
\begin{split}
    \mathrm{d}\mathbf{x}_t &= \mathbf{f}(\mathbf{x}_t,t)\mathrm{d}t - \frac{1}{2}\nabla\cdot[\mathbf{G}(\mathbf{x}_t, t)\mathbf{G}(\mathbf{x}_t, t)^\top]\mathrm{d}t \\
    & -\frac{1}{2} \mathbf{G}(\mathbf{x}_t, t)\mathbf{G}(\mathbf{x}_t, t)^\top\nabla_x\textrm{log }p_t(\mathbf{x}_t)\mathrm{d}t + \mathbf{G}(\mathbf{x}_t,t)\mathrm{d}\mathbf{\bar{w}}
\end{split}
\end{equation}

where time now flows backwards from $t=1$ to $t=0$ and $\nabla\cdot \mathbf{F}(\mathbf{x}) := [\nabla\cdot \mathbf{f}_1(\mathbf{x}), \cdots,  \nabla\cdot \mathbf{f}_d(\mathbf{x})]^\top$
for a matrix-valued function $\mathbf{F}(\mathbf{x}) = [\mathbf{f}_1(\mathbf{x}), \cdots, \mathbf{f}_d(\mathbf{x})]^\top$. The goal of diffusion models is to approximate the score $\nabla_x\textrm{log }p_t(\mathbf{x}_t)$ and use the reverse SDE to sample from the generative model. The score can be approximated by $\mathbf{s}_{\theta}(\mathbf{x}_t, t)$ which provides the following objective

\begin{equation} \label{eqn:score-optim}
\begin{split}
    \theta^{*} &= \textrm{argmin}_{\theta} \mathbb{E}_{t\sim U[0,1]} \mathbb{E}_{\mathbf{x}_0 \sim p_0(\mathbf{x})} \mathbb{E}_{\mathbf{x}_t\sim p_{0t}(\mathbf{x}_t | \mathbf{x}_0)} \\
    & \lambda(t) \left[\Vert \mathbf{s}_{\theta}(\mathbf{x}_t, t) - \nabla_{\mathbf{x}_t}\textrm{log }p_{0t}(\mathbf{x}_t | \mathbf{x}_0) \Vert^2_2\right]
\end{split}
\end{equation}

where $\lambda(t)$ is a weighting function and $p_{st}(\mathbf{x}_t | \mathbf{x}_s)$ is the transition kernel from $x(s)$ to $x(t)$. We note that a number of other objectives can be used to learn the score function \cite{song2021scorebased}. A common practice when using diffusion models is to discretize time into uniform steps \cite{diff_ho}.

\section{Method}
\subsection{The Logistic-Normal Distribution on the Probability Simplex}
Recall the definition of the probability simplex $\mathcal{S}^{k}$. We interpret points in the probability simplex as probability distributions over $k$ categories.

The logistic-normal distribution is an example of a probability distribution over the probability simplex. It is defined as the probability distribution of a random variable whose multinomial logit is a normal distribution, (or equivalently it is the distribution of the softmax function applied to a Gaussian, see \eqref{eq:softmax}). The probability density function of the logistic normal is

\begin{equation} \label{eqn:score-optim}
\begin{split}
    & p(\mathbf{x}; \mathbf{\mu}, \mathbf{\Sigma}) = \frac{1}{\vert (2\pi)^{d-1}\mathbf{\Sigma} \vert} \frac{1}{\prod_{i=1}^{d}\mathbf{x}_i} \\
    &\mathrm{exp }\left(-\frac{1}{2}\left[\log\left(\frac{\bar{\mathbf{x}}_d}{\mathbf{x}_d}\right)-\mathbf{\mu}\right]^\top \mathbf{\Sigma}^{-1} \left[\log\left(\frac{\bar{\mathbf{x}}_d}{\mathbf{x}_d}\right)-\mathbf{\mu}\right]\right)
\end{split}
\end{equation}

where $\mathbf{x}\in\mathcal{S}^d$ and $\bar{\mathbf{x}}_d = [x_1,\dots,x_{d-1}]^\top$. In the $d=2$ dimensional case, the distribution can be understood as mapping a Gaussian distribution on $\mathbb{R}$ to $[0,1]$ via the sigmoid function.

\begin{figure}[ht]
\centering
    \includegraphics[width=230px, keepaspectratio]{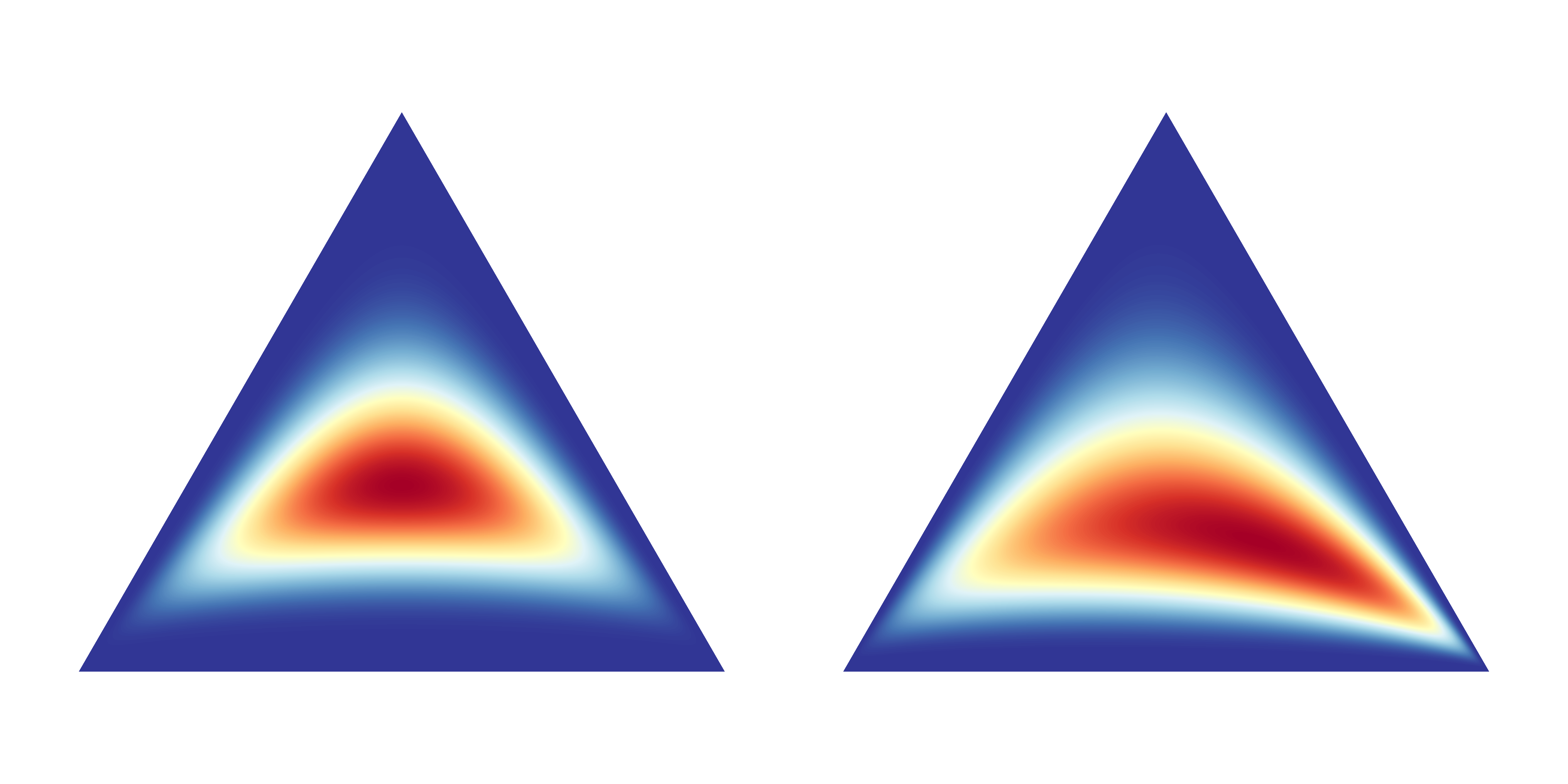}
    \caption{Examples of the Logistic-Normal distribution (PDF values) on $\mathcal{S}^3$ with parameters $\mu = [0,0],~[0.2,0.35]$ and $\sigma = [0.5,0,5],~[0.6,0.8]$ respectively.}
\label{fig:score_dist}
\end{figure}

To constructively sample from this distrubution, we map a point $\mathbf{y}\in\mathbb{R}^{d-1}$ to a point in the probability simplex $\mathbf{x}\in\mathcal{S}^d$ using the additive logistic transformation $\sigma : \mathbb{R}^{d-1} \to \mathcal{S}^d$ defined by  

\begin{equation}\label{eq:softmax}
    \mathbf{x}_i = \sigma_i(\mathbf{y}) := 
\begin{dcases}
    \frac{e^{{\mathbf{y}}_i}}{1+\sum_{k=1}^{d-1}e^{{\mathbf{y}}_k}},& \text{if } i\in \{1,\dots,d-1\}\\
    \frac{1}{1+\sum_{k=1}^{d-1}e^{{\mathbf{y}}_k}},& \text{if } i=d
\end{dcases}
\end{equation}

Where we note that $1 - \sum_{i=1}^{d-1}\mathbf{x}_i = (1+\sum_{k=1}^{d-1}e^{\mathbf{y}_k})^{-1}$. Conversely, the unique inverse map from $\mathcal{S}^d$ to $\mathbb{R}^{d-1}$ is  
$$\mathbf{y}_i = \textrm{log }\left[\frac{\mathbf{x}_i}{\mathbf{x}_d}\right], i\in \{1,\dots,d-1\}.$$


\subsection{The Ornstein-Unlenbeck Process}
The Ornstein-Unlenbeck (OU) process is a real-valued stochastic process used in financial mathematics and physical sciences. Originally, it was developed to model the velocity of a Brownian particle under the force of friction. The process can be described by the following stochastic differential equation:

$$\mathrm{d}\mathbf{Y}_t = -\theta \mathbf{Y}_t dt + \sigma \mathrm{d}\mathbf{W}_t$$

where $\theta > 0$ and $\sigma > 0$ are parameters and $\mathbf{W}_t$ is the WWiener process. The distribution at time $t$ of the process is given by a normal distribution

$$\mathbf{Y}_t \stackrel{d}{=} \mathcal{N}\left(\mathbf{Y}_0e^{-\theta t}, \frac{1}{2\theta}\left(1-e^{-2\theta t}  \right) \mathbf{I} \right).$$

In the limit as $t \rightarrow \infty$ the process has a distribution of $\mathcal{N}\left(0, \frac{1}{2\theta}\right)$, meaning that $\theta$ uniquely determines the limiting distribution.

\subsection{Diffusion on the Probability Simplex}
Our main contribution is a novel diffusion process that operates on the probability simplex. Our method works by first defining the forward process by using the additive logistic transformation from equation \ref{eq:softmax} to map an OU process from $\mathbb{R}^{d}$ to $\mathcal{S}^d$.

$$\mathbf{X}_t = \sigma(\mathbf{Y}_t)$$

In our case we are able to get an exact solution for $\mathbf{S}_t$ by pushing forward the solution of the OU process, meaning that $\mathbf{X}_t \sim \sigma\left(\mathcal{N}\left(\mathbf{Y}_0e^{-\theta t}, \frac{1}{2\theta}\left(1-e^{-2\theta t}\right) \right)\right)$. In other words, at each point $t$ we have a closed form representation of the transition kernel $p_{t0}(\mathbf{x}_t|\mathbf{x}_0)$ which is a logistic Gaussian distribution that we can efficiently sample from. Moreover, one can obtain the SDE for $\mathbf{X}_t$ by applying Ito's lemma to the SDE for $\mathbf{Y}_t$. Carrying this out (see appendix \ref{ssec:ito_deriv}) gives

\begin{equation}
d\mathbf{X}_t = f(\mathbf{X}_t,t)d t + \mathbf{G}(\mathbf{X}_t,t) d \mathbf{W}_t
\end{equation}
%
%
%
%
where the diffusion coefficient matrix $\mathbf{G}$ can be written as:
\[
    \mathbf{G}_{ij}(\mathbf{x},t) = 
\begin{dcases}
    \mathbf{x}_i(1 - \mathbf{x}_i), & i=j \\
    -\mathbf{x}_i\mathbf{x}_j, & i \neq j
\end{dcases}
\]
and the drift term $f$ can be written as: 
$$\mathbf{f}_{i}(\mathbf{x},t) = -\theta\mathbf{x}_i\left[(1-\mathbf{x}_i)\textbf{a}_i + \sum_{j\neq i}\mathbf{x}_j \textbf{a}_j \right]$$

where $\textbf{a}_j = \textbf{x}_j + \frac{1}{2}(1-2\mathbf{x}_j)$.

In order to train the score-matching model, we must also have a closed form solution of $\nabla_x\textrm{log }p(\mathbf{x})_i$, which we show in Appendix \ref{ssec:score_deriv}. The results of the derivation is that the score of the logistic-normal distribution is

\begin{equation} \label{eqn:score-optim}
\begin{split}
    \nabla_x\textrm{log }p(\mathbf{x})_i = -&\frac{1}{v}\left(\frac{1}{x_d}\sum_{k=1}^{d-1}\sigma^\mu_k(\mathbf{x}) + \frac{1}{x_d} \sigma^\mu_i(\mathbf{x}) \right) \\
    & + \frac{\mathbf{x}_i - \mathbf{x}_d}{x_ix_d}
\end{split}
\end{equation}

where we write $\sigma^\mu_k(\mathbf{x}) = \log\left[\frac{\mathbf{x}_i}{\mathbf{x}_d}\right] - \mu$. Finally, the calculation for deriving $\nabla\cdot[\mathbf{G}(\mathbf{x}_t, t)\mathbf{G}(\mathbf{x}_t, t)^\top]$ is performed in Appendix \ref{ssec:diff-matrix-div}.

\subsection{Implementation Considerations}
An example application of this model is for modelling discrete data. A dataset with $k$ different categories, can naturally be modelled with the simplex in $\mathcal{S}^k$. The data distribution could then be represented as a linear combination of Dirac delta functions centered at the corners of the simplex at $t=0$. In other words, each data sample would correspond to a one-hot vector. In practice we relax this condition such that at the beginning of the forward process, data samples are mapped to vectors $\mathbf{x} = [\alpha, \beta, \cdots, \beta]^\top$, where $\beta = \frac{1-\alpha}{d-2}$. For example, a reasonable choice of $\alpha$ would be $0.9$ if $k=6$.

During the optimization process, the score suffers from numerical instability in perimeter regions on the simplex. Furthermore, the region around the perimeter increases as the dimension of the simplex $d$ grows. To deal with this problem, we notice that we directly predict the term $-\frac{1}{2} \mathbf{G}(\mathbf{x}, t)\mathbf{G}(\mathbf{x}, t)^\top\nabla_x\textrm{log }p_t(\mathbf{x})$ from the reverse diffusion SDE.

\begin{figure}[H]
\centering
    \includegraphics[width=230px, keepaspectratio]{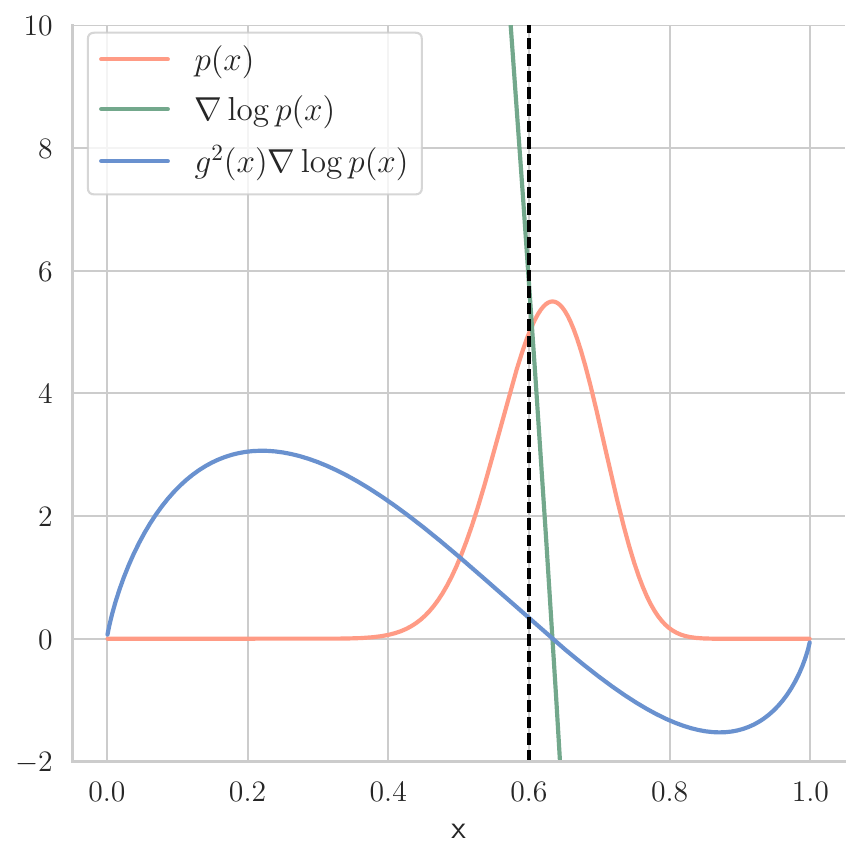}
    \caption{A comparison between the regular score, $\nabla_x\textrm{log }p_t(x)$, and the reverse SDE term, $g^2(x,t)\nabla_x\textrm{log }p_t(x)$, in the one-dimensional case. The reverse SDE term is bounded at the border of the interval $[0,1]$, unlike the score. The PDF of the logistic-normal distribution is plotted for clarity, along with a dotted line around the score for visual clarity.}
\label{fig:score_dist}
\end{figure}
\captionsetup{belowskip=0pt}

\section{Results}
We present initial results of the Simplex Diffusion model using the MNIST dataset. We create a discrete version of the dataset which maps the pixel values that are typically in $[0, 1, \cdots, 255]$ to $[0,1,2]$ for a total of $k=3$ unique categories. In our experiments we use the following parameters: $\theta = 20$, $\alpha=0.9$ and $t\in[0.01, 0.25]$. We parameterize the score function by a U-Net \cite{unet} model with 35 million parameters.

When samples are generated, they must be converted from vectors on the probability simplex, to one of $k$ discrete categories. We choose to take the argmax of the sampled vectors to convert from points on the simplex to discrete categories. Qualitative results from this initial experiment can be found in Figure \ref{fig:score_dist}.

\begin{figure}[H]
\centering
    \includegraphics[width=220px, keepaspectratio]{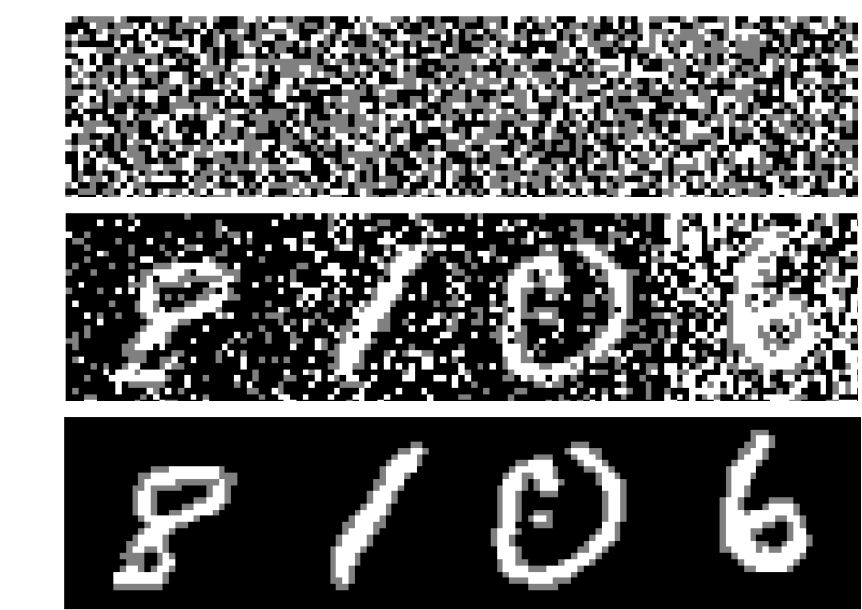}
    \caption{Random samples from a Simplex Diffusion model. Samples are taken at the beginning, middle and end of the reverse process and correspond to the top middle and bottom row respectively. Sampling is done with $T=1000$ denoising steps}
\label{fig:score_dist}
\end{figure}

\section{Discussion}
Our methodology is related to recent works extending diffusion to the bounded domains of the probability simplex and the unit cube. In this section we compare these methodologies with our proposed model to highlight important differences.

\subsection{Simplex Diffusion}
Categorical SDEs with Simplex Diffusion \citep{simplex} use a diffusion process of Gamma random variables to sample from a Dirichlet distribution over the simplex. The Dirichlet distribution is an appealing choice as it is the conjugate prior of the categorical distribution . The forward process used is the Cox-Ingersoll-Ross process, which is defined by the SDE $\mathrm{d}\theta = b(a-\theta)\mathrm{d}t + \sigma\sqrt{2b\theta} dw$, where $\theta(t=0)\geq0$ and $a,b,\sigma > 0$. A drawback of this approach is that while the process has a limiting distribution that is Dirichlet, this is not the case during the transient regime of the process dynamics. 

Our proposed diffusion with the OU process and the Logit-Normal distribution remains a Logistic-Normal distribution throughout the diffusion process due to the correspondence between diffusion spaces in $\mathbb{R}^d$ and $\mathcal{S}^d$ via Ito's lemma.

\subsection{Unit-Cube Diffusion}
Reflected Diffusion \cite{reflected} is a method of performing diffusion on the unit cube $[0,1]^d$ that is motivated by applications to pixel-based diffusion models. When image based diffusion models are used with Gaussian noise, sampling errors often compound and result in pixel values that are outside the valid data range of the unit cube. To mitigate this problem, thresholding is often performed to keep generated images to reasonable values via knowledge of the data distribution constraints \cite{diff_ho} \cite{dhariwal2021diffusion}. While thresholding is popular in many image based diffusion models, it is theoretically unsound as there is a disconnect between the training and generative processes. The authors address this problem by using a reflected diffusion process that reflects particle trajectories into the interior of a data domain $\Omega$ that would normally extend outside the domain.

An interesting property of our Simplex Diffusion Model is that it can be naturally extended to higher dimensions by performing diffusion on the unit cube. By taking the product of $d$ one-dimensional processes that we have developed, we create a diffusion process that is contained to the unit cube. A drawback of the Reflected Diffusion approach is that the resulting score from the forward process cannot be written in closed form. The authors use a combination of two approximations to apply their model in practice. On the other hand, our method maintains an closed form score function that is easy to implement.

\section{Conclusion}
We introduce a novel method to perform diffusion on the probability simplex and the unit cube. In both cases our method allows for an exact solution for the SDE dynamics, and fits into the common diffusion training paradigm.

Future work involves testing the method on more complex datasets and evaluating the properties on the categorical distribution. For example, if the entropy can be utilized as a natural notation of aleatoric uncertainty over generated values.

\nocite{langley00}

\bibliography{manuscript}
\bibliographystyle{icml2023}

\newpage
\appendix
\onecolumn
\section{Mathematical calculations}
\subsection{Score Derivation} \label{ssec:score_deriv}
We want to calculate $\nabla_\mathbf{x}\textrm{log }p(\mathbf{x})$ where
 
$$\textrm{log }p(\mathbf{x}) = -\textrm{log }[Z] - \textrm{log }\left[\prod_{i-1}^{d}x_i\right] -\frac{1}{2v}\bigg\Vert\textrm{log }\left[\frac{\bar{\mathbf{x}}_d}{\mathbf{x}_d} \right] - \mu \bigg\Vert_2^2$$ 

We first find the gradient of second term, given that the log normalizing constant doesn't have a gradient. 

$$
\begin{aligned}
    \alpha &:= -\nabla_\mathbf{x}\textrm{log }\left[\prod_{i=1}^{d}\mathbf{x}_i \right] \\ 
    \alpha_i &= -\frac{\partial}{\partial \mathbf{x}_i}\left(\sum_{i=1}^{d-1}\textrm{log }[\mathbf{x}_i] + \textrm{log }\left[a - \sum_{k=1}^{d-1}\mathbf{x}_k \right]\right) \\
    &= -\frac{1}{\mathbf{x}_i} + \frac{1}{a - \sum_{k=1}^{d-1}\mathbf{x}_k} \\
    &= \frac{1}{\mathbf{x}_d} - \frac{1}{\mathbf{x}_i} \\
    &= \frac{\mathbf{x}_i - \mathbf{x}_d}{\mathbf{x}_i\mathbf{x}_d} \\
\end{aligned}
$$

Next, we deal with the exponential term:

$$
\begin{aligned}
    \beta &:= -\frac{1}{2v}\nabla_\mathbf{x} \bigg\Vert\textrm{log }\left[\frac{\bar{\mathbf{x}}_d}{x_d} \right] - \mu \bigg\Vert_2^2 \\ 
    \beta_i &= -\frac{1}{2v}\frac{\partial}{\partial \mathbf{x}_i}\left(\sum_{k=1}^{d-1}\left(\textrm{log }\left[\frac{\mathbf{x}_k}{\mathbf{x}_d} \right]-\mu \right)^2 \right) \\
    &= -\frac{1}{2v}\sum_{k=1}^{d-1}\left(\frac{\partial}{\partial u}u^2\frac{\partial}{\partial \mathbf{x}_i}u \right), u = \textrm{log }\left[\frac{\mathbf{x}_k}{\mathbf{x}_d} \right]-\mu \\
\end{aligned}
$$

Working with $\kappa := \frac{\partial}{\partial u}u^2\frac{\partial}{\partial \mathbf{x}_i}u$ we get

$$
\begin{aligned}
    \kappa &:= \frac{\partial}{\partial u}u^2\frac{\partial}{\partial \mathbf{x}_i}u \\
    &= 2u \left(\frac{\partial}{\partial \mathbf{x}_i}\textrm{log }[\mathbf{x}_k] - \frac{\partial}{\partial \mathbf{x}_i}\textrm{log} \left[a - \sum_{k=1}^{d-1}\mathbf{x}_k \right]\right) \\
    &= 2u \left(\delta_{ik}\frac{1}{\mathbf{x}_i} + \frac{1}{\mathbf{x}_d}\right) \\
\end{aligned}
$$

Combining terms again we get:

$$
\begin{aligned}
    \beta_i &= -\frac{1}{v}\sum_{k=1}^{d-1}\left(\delta_{ik}\frac{1}{\mathbf{x}_i} + \frac{1}{\mathbf{x}_d}\right)\left(\textrm{log }\left[\frac{\bar{\mathbf{x}}_d}{\mathbf{x}_d} \right]-\mu\right) \\
    &= -\frac{1}{v\mathbf{x}_d}\sum_{k=1}^{d-1}\left(\textrm{log}\left[\frac{\mathbf{x}_k}{\mathbf{x}_d}\right]-\mu \right) - \frac{1}{v\mathbf{x}_i}\left(\textrm{log}\left[\frac{\mathbf{x}_i}{\mathbf{x}_d}\right]-\mu \right) \\
    &= -\frac{1}{v\mathbf{x}_d}\sum_{k=1}^{d-1}\gamma_\mu^k(\mathbf{x}) - \frac{1}{v\mathbf{x}_i}\gamma_\mu^i(\mathbf{x}) \\
\end{aligned}
$$

where we write $\gamma_\mu^i(\mathbf{x}) = \textrm{log}\left[\frac{\mathbf{x}_i}{\mathbf{x}_d}\right]-\mu$

For the final results, we must combine the $\alpha$ and $\beta$ terms together to get:

$$
\begin{aligned}
    \nabla_\mathbf{x}\textrm{log }p_a(\mathbf{x})_i = -\frac{1}{v\mathbf{x}_d}\sum_{k=1}^{d-1}\gamma_\mu^k(\mathbf{x}) - \frac{1}{v\mathbf{x}_i}\gamma_\mu^i(\mathbf{x}) + \frac{\mathbf{x}_i - \mathbf{x}_d}{\mathbf{x}_i\mathbf{x}_d} \\
\end{aligned}
$$

\subsection{Sampling and Ito's Lemma} \label{ssec:ito_deriv}

We are working with an OU process of the following form:

$$\mathrm{d}\mathbf{Y}_t = -\theta\mathbf{Y}_t \mathrm{d}t + \mathrm{d}\mathbf{B}_t$$

with a corresponding process on the simplex:

$$\mathbf{X}_t = \sigma(\mathbf{Y}_t)$$

To keep this section self-contained the definition of $\sigma$ is:

$$\sigma_i(\mathbf{y}) = \frac{e^{\mathbf{y}_i}}{1+\sum_{k=1}^{d-1}e^{\mathbf{y}_k}}, i\in \{1,\dots,d-1\}$$

We must write $\mathbf{X}_t$ in a form where $\mathbf{X}_t = \mathbf{f}(\mathbf{X},t)dt + \mathbf{G}(\mathbf{X},t)dB_t$. This can be done via Ito's Lemma:

$$d\mathbf{X}_i = -\theta(\nabla_X\sigma_i(\mathbf{X}))^\top \mathbf{X}dt + \frac{1}{2}\textrm{Tr}[H_{X}\sigma_i(\mathbf{X})]dt + \nabla_X\sigma_i(\mathbf{X})^\top d\mathbf{B}$$

Where $H_X$ is the Hessian matrix and we drop the time dependence of $\mathbf{S}_t$ and $\mathbf{X}_t$ for notational simplicity. First we deal with the gradient term of the equation. We will use $\gamma(\mathbf{X}) = 1 + \sum_{k=1}^{d-1}e^{\mathbf{X}_k}$ to keep notation smaller.

$$
\begin{aligned}
    \nabla_X\sigma_i(\mathbf{X}) = \mathbf{G} &= \nabla_X\frac{e^{\mathbf{X}_i}}{\gamma(\mathbf{X})} \\
    g_j &= \frac{\partial}{\partial \mathbf{X}_{j}} \frac{e^{\mathbf{X}_i}}{\gamma(\mathbf{X})}
\end{aligned}
$$

We deal with the case when when $j=i$ below
$$
\begin{aligned}
    \mathbf{G}_i &= \frac{\partial}{\partial \mathbf{X}_{i}} \frac{e^{\mathbf{X}_i}}{\gamma(\mathbf{X})} \\
    &= \gamma(\mathbf{X})^{-2}\left[\gamma(\mathbf{X})\frac{\partial}{\partial \mathbf{X}_i} e^{\mathbf{X}_i} - e^{\mathbf{X}_i}\frac{\partial}{\partial \mathbf{X}_i}\gamma(\mathbf{X})\right] \\ 
    &= \gamma(\mathbf{X})^{-2}\left[e^{\mathbf{X}_i}\gamma(\mathbf{X}) - e^{2\mathbf{X}_i}\right] \\ 
    &= \sigma_i(\mathbf{X})\gamma(\mathbf{X})^{-1}[\gamma(\mathbf{X}) - e^{\mathbf{X}_i}] \\
    &= \sigma_i(\mathbf{X})(1-\sigma_i(\mathbf{X})) \\
\end{aligned}
$$

and the case when $j\neq i$:
$$
\begin{aligned}
    \mathbf{G}_j &= \frac{\partial}{\partial \mathbf{X}_{j}} \frac{ae^{\mathbf{X}_i}}{\gamma(\mathbf{X})} \\
    &= -\frac{e^{\mathbf{X}_i}e^{\mathbf{X}_j}}{\gamma(X)^2} \\
    &= -\sigma_i(\mathbf{X})\sigma_j(\mathbf{X}) \\
\end{aligned}
$$

Next we deal with the trace Hessian term:

$$\textrm{Tr}[H_{\mathbf{X}}\sigma^a_i(\mathbf{X})] = \sum_{j=1}^{d-1}\frac{\partial^2}{\partial \mathbf{X}_j^2} \sigma^a_i(\mathbf{X})$$

which again can be split into two cases. First we deal with the case when $j=i$

$$
\begin{aligned}
    \frac{\partial^2}{\partial X_i^2} \sigma^a_i(\mathbf{X}) &= a\frac{\partial}{\partial X_i} \sigma_i(\mathbf{X})(1-\sigma_i(\mathbf{X})) \\
    &= a\sigma_i(\mathbf{X})(1-\sigma_i(\mathbf{X}))(1-2\sigma_i(\mathbf{X})) \\
\end{aligned}
$$

Then the case where $j\neq i$

$$
\begin{aligned}
    \frac{\partial^2}{\partial \mathbf{X}_j^2} \sigma^a_i(\mathbf{X}) &= -\frac{\partial}{\partial \mathbf{X}_j} \sigma_i(\mathbf{X})\sigma_j(\mathbf{X})\\
    &= -\sigma_i(\mathbf{X})\sigma_j(\mathbf{X})(1 - 2\sigma_j(\mathbf{X}))
\end{aligned}
$$

In summary the diffusion and drift terms are:

\[
    \mathbf{G}_{ij}(\mathbf{x},t) = 
\begin{dcases}
    \mathbf{x}_i(1 - \mathbf{x}_i), & i=j \\
    -\mathbf{x}_i\mathbf{x}_j, & i \neq j
\end{dcases}
\]
$$\mathbf{f}_{i}(\mathbf{x},t) = -\theta\mathbf{x}_i\left[(1-\mathbf{x}_i)\textbf{a}_i + \sum_{j\neq i}\mathbf{x}_j \textbf{a}_j \right]$$

where $\textbf{a}_j = \textbf{x}_j + \frac{1}{2}(1-2\mathbf{x}_j)$

\subsection{Diffusion Matrix Divergence} \label{ssec:diff-matrix-div}

Suppose we have $\mathbf{x}$, which is some position on the probability simplex, and $\mathbf{G}$ from above in \ref{ssec:ito_deriv}. We want $\nabla\cdot  \mathbf{G}(\mathbf{x}) \mathbf{G}(\mathbf{x})^\top$ where the definition of the matrix divergence over matrix valued function $\mathbf{F}$ is (defined similarly as Appendix A. in \citet{song2021scorebased}):

$$
\nabla\cdot \mathbf{F}(\mathbf{x}) := [\nabla\cdot \mathbf{f}^1(\mathbf{x}), \nabla\cdot \mathbf{f}^2(\mathbf{x}), ...]^\top
$$

where $\mathbf{F}(\mathbf{x}) = [f_1(\mathbf{x}), f_2(\mathbf{x}), ...]^\top$. To further clarify some terms, we start with $\mathbf{G}(\mathbf{x}) = \mathbf{G}(\mathbf{x})^\top$, which gives us the Hessian as $\mathbf{H} = \mathbf{G}(\mathbf{x})\mathbf{G}(\mathbf{x})^\top$. Equivalently, the Hessian in coordinate-wise notation is: 

$$ \mathbf{H}_{ij} = \sum_k \mathbf{G}_{ik}(\mathbf{x})\mathbf{G}_{kj}(\mathbf{x}) $$

We will then being our derivation by analyzing $\mathbf{G}(.)$, which decomposes into two cases:

$$
\mathbf{G}_{ij}(\mathbf{x})=\begin{cases}
\mathbf{x}_i(1-\mathbf{x}_i),& \text{when} ~ i=j \\
-\mathbf{x}_i\mathbf{x}_j,& \text{when} ~ i\neq j
\end{cases}
$$

Starting with case 1, $i=j$:
$$
\begin{aligned}
    \mathbf{H}_{ii} &= \sum_k \mathbf{G}_{ik}(\mathbf{x})\mathbf{G}_{ki}(\mathbf{x}) \\
    &= \mathbf{x}_i^2(1-\mathbf{x}_i)^2 + \mathbf{x}_i^2\sum_{k\neq i}\mathbf{x}_k^2 \\
    &= \mathbf{x}_i^2\left((1-\mathbf{x}_i)^2 + \sum_{k\neq i}\mathbf{x}_k^2\right)
\end{aligned}
$$

Then for case 2, $i\neq j$:

$$
\begin{aligned}
    \mathbf{H}_{ij} &= \sum_k \mathbf{G}_{ik}(\mathbf{x})\mathbf{G}_{kj}(\mathbf{x}) \\
    &= -\mathbf{x}_i(1-\mathbf{x}_i)\mathbf{x}_i\mathbf{x}_j -\mathbf{x}_j(1-\mathbf{x}_j)\mathbf{x}_j\mathbf{x}_i + \mathbf{x}_i\mathbf{x}_j\sum_{k\neq i,j}\mathbf{x}_k^2 \\
    &= -\mathbf{x}_i^2\mathbf{x}_j(1-\mathbf{x}_i) -\mathbf{x}_j^2\mathbf{x}_i(1-\mathbf{x}_j) + \mathbf{x}_i\mathbf{x}_j\sum_{k\neq i,j}\mathbf{x}_k^2 \\
    &= -\mathbf{x}_i\mathbf{x}_j \left(\mathbf{x}_i(1-\mathbf{x}_i) +\mathbf{x}_j(1-\mathbf{x}_j) -\sum_{k\neq i,j}\mathbf{x}_k^2 \right)
\end{aligned}
$$

Now, let $\mathbf{d}$ be the divergence of $\mathbf{H}$ as defined at the start of this derivation:

$$
\mathbf{d} := \nabla\cdot \mathbf{H} = [\nabla\cdot \mathbf{h}_1(\mathbf{x}), \nabla\cdot \mathbf{h}_2(\mathbf{x}), ...]^\top
$$

where $\mathbf{h}_i$ is a \emph{row} vector. Then continuing, we have that $\mathbf{d}_i$ is given as the following summation:

$$
\begin{aligned}
    \mathbf{d}_i &= \nabla\cdot \mathbf{h}_i \\
    &= \sum_k \frac{\partial}{\partial \mathbf{x}_k}\mathbf{h}_{ik} \\
\end{aligned}
$$

From the summation, we again have two cases, first when $k=i$ and second when $k \neq i$. Starting with case 1, i.e., when $k=i$, we first construct a "helper" function $a(\mathbf{x}_i)$ such that:
$$
\begin{cases}
a(\mathbf{x}_i) = (1-\mathbf{x}_i)^2 + \sum_{k\neq i}\mathbf{x}_k^2 \\
\frac{\partial}{\partial \mathbf{x}_i} a(\mathbf{x}_i) = -2(1-\mathbf{x}_i)
\end{cases}
$$

Then using $a(\mathbf{x}_i)$ we can express $\frac{\partial}{\partial \mathbf{x}_i}\mathbf{h}_{ii}$ as:

$$
\begin{aligned}
    \frac{\partial}{\partial \mathbf{x}_i}\mathbf{h}_{ii} &= \frac{\partial}{\partial \mathbf{x}_i} \mathbf{x}_i^2a(\mathbf{x}_i) \\
    &= 2\mathbf{x}_i \left((1-\mathbf{x}_i)^2 + \sum_{k\neq i}\mathbf{x}_k^2 \right) - 2(1-\mathbf{x}_i)\mathbf{x}_i^2
\end{aligned}
$$

Now, examining case 2, i.e., when $k \neq i$, we can again define another "helper" function $b(\mathbf{x}_i)$ such that:

$$
\begin{cases}
b(\mathbf{x}_i) = \mathbf{x}_i(1-\mathbf{x}_i) +\mathbf{x}_j(1-\mathbf{x}_j) -\sum_{k\neq i,j}\mathbf{x}_k^2 \\
\frac{\partial}{\partial \mathbf{x}_i} b(\mathbf{x}_i) = (1-\mathbf{x}_i) - \mathbf{x}_i = 1- 2\mathbf{x}_i
\end{cases}
$$

Using $b(\mathbf{x}_i)$ leads us to the following for $\frac{\partial}{\partial \mathbf{x}_i}\mathbf{h}_{ij}$:

$$
\begin{aligned}
    \frac{\partial}{\partial \mathbf{x}_i}\mathbf{h}_{ij} &= -\frac{\partial}{\partial \mathbf{x}_i} \mathbf{x}_i\mathbf{x}_jb(\mathbf{x}_i) \\
    &= -\mathbf{x}_j \left(\mathbf{x}_i(1-\mathbf{x}_i) +\mathbf{x}_j(1-\mathbf{x}_j) -\sum_{k\neq i,j}\mathbf{x}_k^2 \right) -\mathbf{x}_i\mathbf{x}_j(1- 2\mathbf{x}_i) 
\end{aligned}
$$

Finally, we are left to combine the previous results in order to derive $\mathbf{d}_i$:

$$
\begin{aligned}
    \mathbf{d}_i &= \nabla\cdot \mathbf{h}_i \\
    &= \sum_j \frac{\partial}{\partial \mathbf{x}_j}\mathbf{h}_{ij} \\
    &= \frac{\partial}{\partial \mathbf{x}_i}\mathbf{h}_{ii} + \sum_{j\neq i} \frac{\partial}{\partial \mathbf{x}_j}\mathbf{h}_{ij}
\end{aligned}
$$

Also note that we can further expand the above expression to obtain the following:

$$\mathbf{d}_i = 2\mathbf{x}_i \left((1-\mathbf{x}_i)^2 + \sum_{k\neq i}\mathbf{x}_k^2 \right) - 2(1-\mathbf{x}_i)\mathbf{x}_i^2 - \sum_{j\neq i} \left[\mathbf{x}_j \left(\mathbf{x}_i(1-\mathbf{x}_i) +\mathbf{x}_j(1-\mathbf{x}_j) -\sum_{k\neq i,j}\mathbf{x}_k^2 \right) + \mathbf{x}_i\mathbf{x}_j(1- 2\mathbf{x}_i) \right]$$


\end{document}